\documentclass[journal]{IEEEtran}
\usepackage{amsmath}
\usepackage[pdftex]{graphicx}
\usepackage{subfigure}
\usepackage{amsmath}
\usepackage{amssymb}
\usepackage{amsfonts}
\usepackage{lmodern}
\usepackage{float}
\usepackage{bm}
\usepackage{comment}
\usepackage{color}

\usepackage{booktabs}
\usepackage{graphicx}
\usepackage{multirow}
\usepackage{stfloats}
\usepackage{array} 
\usepackage{caption}
\usepackage{algorithm}
\usepackage{algorithmic}

\usepackage[pdftex]{hyperref}

\newcommand\orcidAuthor[1]{
\hspace*{-1mm}\includegraphics[keepaspectratio,width=0.7em]{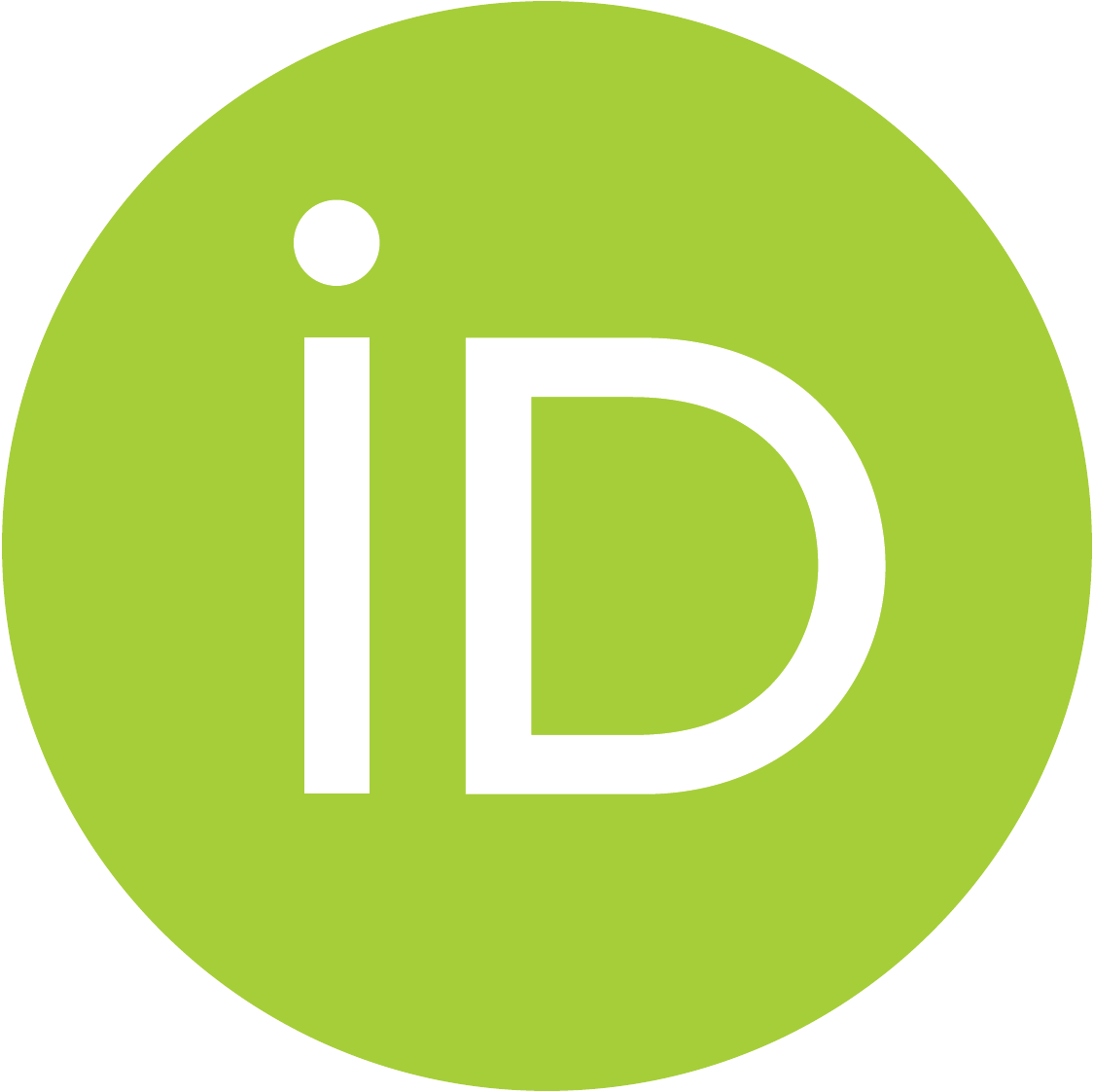}\href{https://orcid.org/#1}
}

\begin{document}
\title{Generative Diffusion Contrastive Network for Multi-View Clustering}

\author{%
Jian~Zhu$^{\dagger}$,~\IEEEmembership{Student Member,~IEEE},
Xin~Zou$^{\dagger}$,
Xi~Wang$^{\dagger}$,
Lei~Liu,~\IEEEmembership{Member,~IEEE},\\
Chang~Tang$^{\ast}$,~\IEEEmembership{Senior Member,~IEEE},
and~Li-Rong~Dai,~\IEEEmembership{Member,~IEEE}\\
\thanks{This work is supported by the National Key Research and Development Program of China (Grant No. 2021ZD0201501), the National Natural Science Foundation of China (No. 22574146, No. 32200860, and No. 62306289).}
\thanks{Jian Zhu is with Zhejiang Lab, China.}
\thanks{Xin~Zou and Xi Wang are from Hong Kong University of Science and Technology, China.}
\thanks{Lei Liu and Li-Rong Dai are with the University of Science and Technology of China, China.}
\thanks{Chang Tang is with Huazhong University of Science and Technology, China.}
\thanks{$\dagger$Contributed equally.} \thanks{$^{\ast}$Corresponding author \{tangchang\}@hust.edu.cn.}
}

%
%

\markboth{IEEE Signal Processing Letters,~Vol.~xx, No.~xx, January~2026}%
{Shell \MakeLowercase{\textit{et al.}}: Bare Demo of IEEEtran.cls for IEEE Journals}

%


\maketitle

\begin{abstract}
 In recent years, Multi-View Clustering (MVC) has been significantly advanced under the influence of deep learning. By integrating heterogeneous data from multiple views, MVC enhances clustering analysis, making multi-view fusion critical to clustering performance. However, there is a problem of low-quality data in multi-view fusion. This problem primarily arises from two reasons: 1) Certain views are contaminated by noisy data. 2) Some views suffer from missing data. This paper proposes a novel Stochastic Generative Diffusion Fusion (SGDF) method to address this problem. SGDF leverages a multiple generative mechanism for the multi-view feature of each sample. It is robust to low-quality data. Building on SGDF, we further present the Generative Diffusion Contrastive Network (GDCN). Extensive experiments show that GDCN achieves the state-of-the-art results in deep MVC tasks. The source code is publicly available at https://github.com/HackerHyper/GDCN.
\end{abstract}

\begin{IEEEkeywords}
Multi-view Clustering, Multi-view Fusion, Diffusion Model, Deep Clustering
\end{IEEEkeywords}

%
\IEEEpeerreviewmaketitle

\section{Introduction}
\IEEEPARstart{W}{ith} the development of artificial intelligence, intelligent systems integrate multiple data sources for analysis, modeling, and decision-making \cite{zheng:82,zhang:83,likun:84,xia:85,ping:86}. For example, autonomous car system aggregates information of data from many cameras to make decisions \cite{chen2024end}. Multi-View Clustering (MVC)~\cite{xiao2024dual} aims to fuse the data of multiple views to discover meaningful groupings, making it crucial to data mining \cite{xu2025hstrans}.

\begin{figure*}
  \centering
  \includegraphics[width=12.5cm]{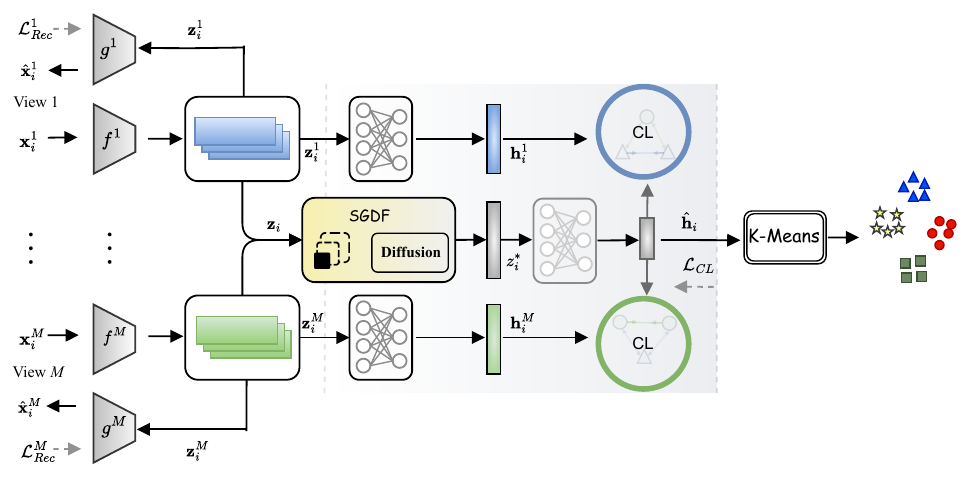}
  \caption{The framework of GDCN. The framework consists of the Autoencoder module, the SGDF module, and the Contrastive Learning (CL) module. The Autoencoder module aims to obtain view-specific representations that effectively reconstruct the original data. The SGDF module performs robust fusion of the multi-view data. Eventually, the Contrastive Learning module is used to get a common representation. K-Means is a module used for clustering.}
  \label{fig:02}
\end{figure*}

In computer vision tasks, deep learning has proven to be quite effective at representing data~\cite{zou2023hierarchical}. Similarly, MVC has experienced tremendous development driven by deep learning~\cite{trosten2021reconsidering, xu2022multi}. These methods leverage a view-specific encoder network to generate effective embeddings, which are then combined from all views for MVC. This leads to the problem of semantic gaps in various view features. To solve the problem, different alignment methods have been proposed. For instance, some approaches utilize KL divergence to align the label or representation distributions across multiple views~\cite{hershey2007approximating}. Recently, many methods have tended to contrastive learning to align representations from multiple views \cite{zhu:81}. Although these methods have made substantial progress in addressing the MVC task, the issue of low-quality data remains. This problem can be attributed to two reasons: 1) Noisy data is present in certain views. 2) Some views have missing data. At present, multi-view fusion methods include summation, concatenation, and attention mechanisms. These methods are sensitive to low-quality data. Therefore, the problem of low-quality data ultimately reduces the performance of MVC.

We propose a Stochastic Generative Diffusion Fusion (SGDF) method to address the problem of low-quality data in multi-view fusion. Inspired by the outstanding features of the diffusion model~\cite{ho2020denoising}, SGDF utilizes a multiple generative mechanism for multi-view data of a sample. It generates multiple features for this sample and then averages them to obtain the fused features. It is a robust method for low-quality data. On the basis of SGDF, we present the Generative Diffusion Contrastive Network (GDCN). It consists of the Autoencoder module, the SGDF module, and the Contrastive Learning module. GDCN first uses the Autoencoder module to obtain view-specific representations that effectively reconstruct the original data. GDCN then utilizes the SGDF module to perform robust fusion of the multi-view data. Finally, GDCN uses the Contrastive Learning module to get a common representation. The contributions of this paper are summarized as follows:
\begin{itemize}
\item We propose a novel SGDF method to solve the problem of low-quality data. SGDF implements the robust fusion of multi-view data through a multiple generative mechanism.
\item Based on SGDF, we present the GDCN for the deep MVC task. It is a simple and effective representation learning method for multi-view data.
\item The comprehensive experiments demonstrate that GDCN achieves the state-of-the-art results on four public datasets.
\end{itemize}

\section{The Proposed Methodology}
\label{section:proposed_method}
We propose an innovative Stochastic Generative Diffusion Fusion (SGDF) method, which aims to solve the problem of low-quality data in multi-view fusion. On the basis of SGDF, we present the Generative Diffusion Contrastive Network (GDCN) for Multi-View Clustering (MVC). As shown in Fig. \ref{fig:02}, GDCN mainly consists of three components: 1) Autoencoder Module, 2) Stochastic Generative Diffusion Fusion, and 3) Contrastive Learning (CL) Module. The multi-view data, which includes $N$ samples with $M$ views, is denoted as $\{\mathbf{X}^{m}=\{{x}_{1}^{m};...;{x}_{N}^{m}\}\in \mathbb{R}^{N \times D_{m}}\}_{m=1}^M$, where $D_m$ is the feature dimension in the $m$-th view. 

\subsection{Autoencoder Module}
We use Autoencoder \cite{song2018self} to extract individual view features. The encoder function is denoted by $f^{m}$ for the $m$-th view. The encoder generates the low-dimensional embedding as follows:
\begin{equation}
\label{eq:encoder}
\begin{aligned}
{z}_{i}^{m}=f^{m}\left({x}_{i}^{m}\right), 
\end{aligned}
\end{equation} 
where ${z}_{i}^{m} \in \mathbb{R}^{d_m}$ is the embedding of the $i$-th sample from the $m$-th view ${x}_{i}^{m}$. $d_m$ is the dimension of the feature.

Using the data representation ${z}_{i}^{m}$, the decoder reconstructs the sample. Let $g^{m}$ denote the decoder function. In the decoder component, ${z}_{i}^{m}$ is decoded to provide the reconstructed sample ${\hat{x}}_{i}^{m}$:
\begin{equation}
\label{eq:decoder}
\begin{aligned}
\hat{{x}}_{i}^{m}=g^{m}\left({z}_{i}^{m}\right).
\end{aligned}
\end{equation}

\subsection{Stochastic Generative Diffusion Fusion} 
\label{sec:model}
All the extracted embeddings $\{z_i^m\}_{m=1}^{M}$ are concatenated as a condition feature:
\begin{equation}
c_i = cat( z_i^1,z_i^2,\dots,z_i^M ),~c_i \in \mathbb{R}^{M d_{m}}, \label{eq:concat}
\end{equation}
where $cat$ denotes a concatenated operator.

\subsubsection{Diffusion Reverse Process}
The reverse process refers to the iterative operation of gradually restoring the original data structure from noisy data. We adopt the sqrt noise scheduling strategy~\cite{li2022diffusion} as follows:
\begin{equation}
\bar{\alpha}_t = 1 - \sqrt{\frac{t}{T} + 10^{-4}}, \label{eq:noise_schedule}
\end{equation}
where $T$ is the total diffusion time steps and $t$ is the current time step. We use a multi-layer perceptron (MLP) as the denoising network to predict the clean data:
\begin{equation}
\mathbf{z}_0 = \text{MLP}\left( \text{cat}(\mathbf{z}_t, c_i) \right), \label{eq:denoise_net}
\end{equation}
where $\mathbf{z}_t$ is the noisy embedding at time step $t$, and $\mathbf{z}_0$ denotes the predicted clean embedding. Following the DDIM sampling strategy, the single-step denoising process transitions from time step $t$ to the next time step $s$ (where $s < t$) as follows:
\begin{equation}
\small
\mathbf{z}_{s} = q(\mathbf{z}_t, c_i, t) = 
\begin{cases} 
\mathbf{z}_0 & t = 0 \\
\sqrt{\alpha_{\text{ratio}}} \cdot \mathbf{z}_t + \sqrt{1 - \alpha_{\text{ratio}} - \sigma^2} \cdot \mathbf{z}_0 & t > 0
\end{cases},
\label{eq:p_sample}
\end{equation}
where $\alpha_{\text{ratio}} = \frac{\bar{\alpha}_s}{\bar{\alpha}_t}$ and $\sigma^2 = \frac{1-\bar{\alpha}_s}{1-\bar{\alpha}_t}$. Here, $s$ denotes the next time step in the reverse process with $s < t$. The sampling proceeds by iteratively decreasing $t$ from near $T$ to $0$.

\subsubsection{Multiple Generative Mechanism}
Multiple Generative Mechanism starts the sampling initialization as follows:
\begin{equation}
\mathbf{z}_T^{(b)} \sim \mathcal{N}(\mathbf{0}, \mathbf{I}_d),~b \in[1,2, \dots, B],
\label{eq:noise_init}
\end{equation}
where $\mathcal{N}$ represents a normal distribution. $\mathbf{I}_d$ denotes a $d$-dimensional identity matrix. $\mathbf{z}_T^{(b)}$ is the initial noise vector for the $b$-th sample in the diffusion process. It serves as the starting point for the reverse diffusion process. $B$ represents sampling times.

We create a time-step sequence used for accelerated sampling during the backward diffusion process. We have set $K$ discrete points.
\begin{equation}
 \tau_k = \left\lfloor T - 1 - k \cdot \frac{T-1}{K-1} \right\rfloor,~k \in [0, 1, \dots, K-1],
\label{eq:time_steps}
\end{equation}
where $k$ is the index of the sampling point. $\tau_k$ is the $k$-th sampling time point, with $\tau_0 = T-1$ and $\tau_{K-1} = 0$. $\lfloor \cdot \rfloor$ is the round down function. The reverse sampling route of the diffusion model progressively denoises from high noise levels to clean data:
\begin{equation}
\mathbf{z}_{\tau_k}^{(b)} = q(\mathbf{z}_{\tau_{k-1}}^{(b)}, c_i, \tau_{k-1}),~k \in [1, 2, \dots, K-1],
\label{eq:iterative_denoising}
\end{equation}
where $q$ is the single-step denoising process defined in Eq.~(\ref{eq:p_sample}). $\mathbf{z}_{\tau_{0}}^{(b)} = \mathbf{z}_T^{(b)}$. We conduct $K-1$ iterations to obtain $\mathbf{z}_{\tau_{K-1}}^{(b)}$. Finally, we average the $B$ generated vectors to obtain the fused feature:
\begin{equation}
\mathbf{z}_i^* = \frac{1}{B} \sum_{b=1}^B \mathbf{z}_{\tau_{K-1}}^{(b)}.
\label{eq:feature_fusion}
\end{equation}

\subsection{Loss Function and Clustering module}
Let $\mathcal{L}_{Rec}$ represent the reconstruction loss. $N$ denotes the number of
samples. The following formula is used to calculate the reconstruction loss:
\begin{equation}
\label{eq:reconstruction loss}
\begin{aligned}
\mathcal{L}_{\mathrm{Rec}}
=&\sum_{m=1}^{M}\sum_{i=1}^{N}\left\|{x}_i^{m}-g^{m}\left({z}_{i}^{m}\right)\right\|_{2}^{2}.
\end{aligned}
\end{equation}

The loss of Contrastive Learning \cite{Yan_2023_CVPR} is calculated as follows:
\begin{equation}
\label{eq:lc}
\begin{aligned}
\mathcal{L}_{\mathrm{CL}}=-\frac{1}{2 N} \sum_{i=1}^{N} \sum_{m=1}^{M} \log \frac{e^{\operatorname{C}\left({\hat{h}}_{i}, {h}_{i}^{m}\right) / \tau}}{\sum_{j=1}^{N} e^{(1-{S}_{ij})\operatorname{C}\left({\hat{h}}_{i}, {h}_{j}^{m}\right) / \tau}-e^{1 / \tau}},
\end{aligned}
\end{equation}
where $\tau$ represents the temperature coefficient. ${S}_{ij}$ is the similarity between the $i$-th sample and the $j$-th sample. $C$ denotes the cosine function. $\hat{h}_{i}$ and ${h}_{i}^{m}$ are generated by $\mathbf{z}_i^*$ and $z^{m}_i$. The total loss is calculated as follows:
\begin{equation}
\label{zong}
{\cal L} = {{\cal L}_{\rm{Rec}}} + {{\cal L}_{\rm{CL}}}.\\
\end{equation}
The training of deep multi-view clustering networks is divided into two stages: the pre-training phase and the fine-tuning phase. $\mathcal{L}_{\mathrm{Rec}}$ is used for the pre-training phase, while ${\cal L}$ is used for the fine-tuning phase. We use the K-Means algorithm for the clustering module \cite{mackay2003information}.

\begin{table}[!t]
\centering
\small 
\setlength{\tabcolsep}{2pt} 
\caption{Description of the multi-view datasets.}
\begin{tabular}{ccccc} 
\toprule
Datasets    & Samples & Views & Clusters & View dimensions   \\ 
\midrule
NGs & 500   & 3     & 5    & [2000, 2000, 2000]    \\
Synthetic3d & 600   & 3     & 3  &  [3, 3, 3]   \\
Caltech5V & 1400     & 5     & 7   &  [40, 254, 1984, 512, 928]    \\
Wikipedia & 693     & 2     & 10   &  [128, 10]    \\
\bottomrule
\end{tabular}
\label{tab:Datasets}
\end{table}

\begin{table*}[ht]
\renewcommand\arraystretch{1}
\small
\setlength{\tabcolsep}{5.5pt}
\setlength{\abovecaptionskip}{0.1cm}  
\centering
\caption{The clustering results on four public datasets. The best results are bolded, and the second-best results are underlined.}
\resizebox{\textwidth}{!}{\begin{tabular}{c|ccc|ccc|ccc|ccc} 
\toprule
Datasets       & \multicolumn{3}{c|}{NGs}                           &\multicolumn{3}{c|}{Synthetic3d}          & \multicolumn{3}{c|}{Caltech5V}                           & \multicolumn{3}{c}{Wikipedia}                           \\ 
\midrule
Metrics       & ACC           & NMI           & PUR           & ACC           & NMI           & PUR           & ACC           & NMI           & PUR           & ACC           & NMI           & PUR                          \\
\midrule
DSMVC \cite{tang:58}    &0.4460 & 0.1166 & 0.4460 &  0.9567 & 0.8267 & 0.9567 & 0.5979 & 0.4400 & 0.6121        &  \underline{0.6118} & 0.5537 & \underline{0.6364} \\
DealMVC  \cite{yang:59}  & 0.5940 & 0.4848 &0.5940    & 0.8983 & 0.6904 &0.8983  &  0.6157 &  0.5081 &0.6157                  &  0.4040 & 0.3707 &0.4040 \\
GCFAggMVC \cite{Yan_2023_CVPR}    &  0.6740&  0.5392 &0.6780  &  0.5750 & 0.3462 &0.6283 &  0.2893 & 0.1264 &0.2979                  &  0.2150 &0.1473 &0.2843         \\
SCMVC  \cite{wu:60}   &  0.8740 & 0.7181 &0.8740         &  0.9417 &  0.7944 & 0.9417    &  0.6086 & 0.4210 & 0.6100                  & 0.4733 & 0.4292  &0.5426                   \\
ACCMVC \cite{yan:72}  & \underline{0.9020}& \underline{0.7546}& \underline{0.9020}     & \underline{0.9600}& \underline{0.8381} &\underline{0.9600}     & \underline{0.6807} &\underline{0.5516} &\underline{0.7129}                  & 0.6075& \underline{0.5621}& 0.6349                   \\ \midrule
GDCN (Ours) & \textbf{0.9800} & \textbf{0.9440} & \textbf{0.9800} &\textbf{0.9717} &\textbf{0.8758} &\textbf{0.9717}  & \textbf{0.7071} &\textbf{0.5777} &\textbf{0.7221}        & \textbf{0.6335} & \textbf{0.5638} & \textbf{0.6522}  \\
\bottomrule
\end{tabular}}
\label{tab:Clustering performance1}
\end{table*}

\section{Experiments}
\subsection{Experimental Settings}
We evaluate the proposed GDCN on four public multi-view datasets \cite{Yan_2023_CVPR} (i.e., NGs, Synthetic3d, Caltech5V, and Wikipedia) with different scales (see Table \ref{tab:Datasets}). Three evaluation metrics (including accuracy (ACC), normalized mutual information (NMI), and Purity (PUR)) are used.

\textbf{Compared methods.} 
To evaluate the effectiveness of the proposed method, we compare the GDCN with five state-of-the-art clustering methods, which are all deep methods (including DSMVC \cite{tang:58}, DealMVC \cite{yang:59}, GCFAggMVC \cite{Yan_2023_CVPR}, SCMVC \cite{wu:60}, and ACCMVC \cite{yan:72}).

\begin{figure}[htbp]
\centering
\subfigure[] { 
\includegraphics[width=2.5in]{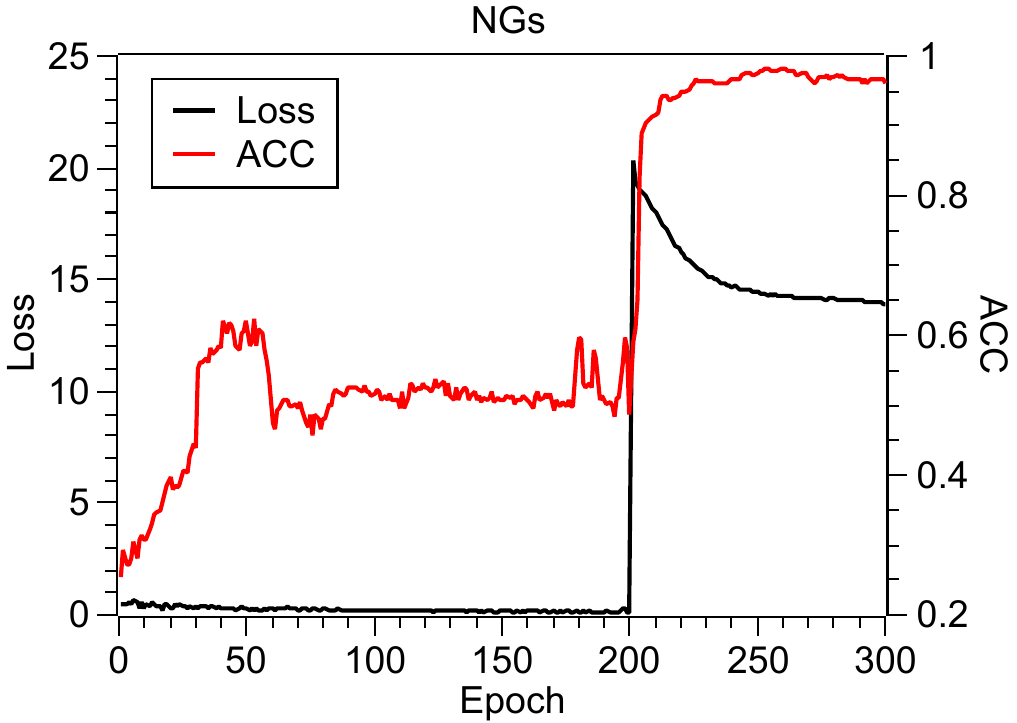}
}\hspace{0.1cm}
\subfigure[] { 
\includegraphics[width=2.5in]{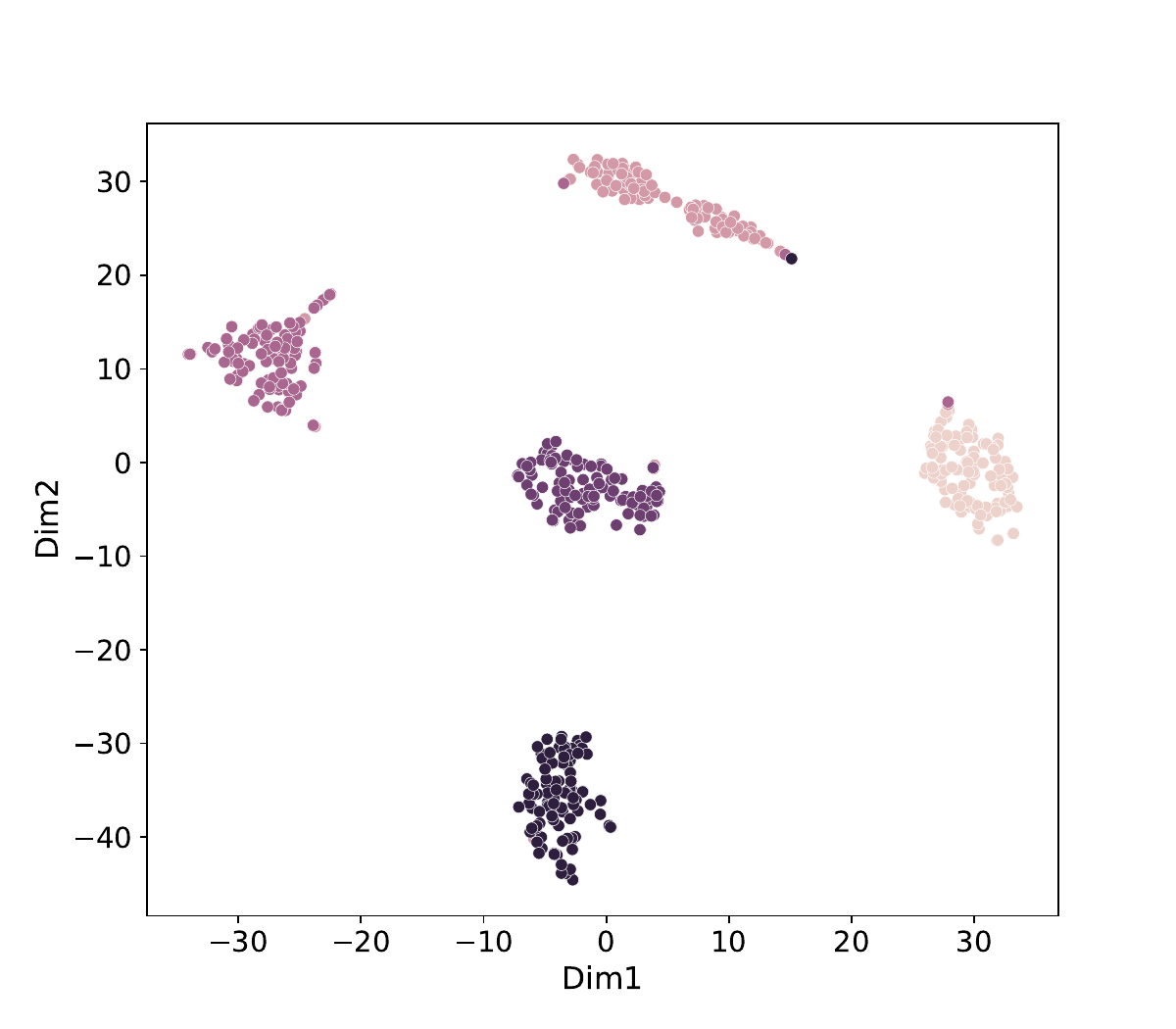}
}
\caption{The convergence analysis and visualization analysis on NGs. (a) The training loss/Acc curves; (b) The t-SNE plot of the common representations $\{\hat{h}_i\}_{i=1}^{N}$.}
\label{fig:Visualization and convergence} 
\end{figure}

\subsection{Experimental comparative results}
The comparative results with five multi-view clustering methods by three evaluation metrics on four benchmark datasets are presented in Table~\ref{tab:Clustering performance1}. 
The results show that GDCN outperforms all the compared baselines by a significant margin on four datasets. Specifically, we obtain the following observations: On the NGs dataset, GDCN outperforms the second-best method ACCMVC by $7.8$ percentage points in ACC. Similarly, on Wikipedia, GDCN performs better than the DSMVC method by $2.17$ percentage points in terms of ACC metrics. GDCN also outperforms the baseline methods significantly in both NMI and PUR metrics.

\subsection{Ablation Study}
\par
We conduct an ablation study to evaluate the two key components of GDCN: Stochastic Generative Diffusion Fusion (SGDF) and Contrastive Learning (CL).
\begin{table}[htb]
\small
\setlength{\tabcolsep}{6pt}
\setlength{\abovecaptionskip}{0.1cm}  
\centering
\caption{Ablation study on diverse datasets.}
\begin{tabular}{ccccc} 
\toprule
Datasets                     & Method      & ACC & NMI  & PUR \\ 
\midrule
\multirow{3}{*}{NGs} & w/o SGDF &  0.8620 & 0.7147  & 0.8620 \\
                             & w/o CL & 0.6180 & 0.4158 & 0.6180 \\
                             & GDCN       & \textbf{0.9800} & \textbf{0.9440} & \textbf{0.9800}  \\
\midrule
\multirow{3}{*}{Synthetic3d}         & w/o SGDF & 0.9600  & 0.8407 & 0.9600 \\
                             & w/o CL & 0.7600 & 0.4500 & 0.7600            \\
                             & GDCN & \textbf{0.9717}  & \textbf{0.8758}& \textbf{0.9717}   \\ 
\midrule
\multirow{3}{*}{Caltech5V} & w/o SGDF &  0.6536 & 0.5127  & 0.6786 \\
                             &  w/o CL & 0.5129 & 0.3299 & 0.5314 \\
                             & GDCN        & \textbf{0.7071} & \textbf{0.5777} & \textbf{0.7221}  \\
\midrule
\multirow{3}{*}{Wikipedia} & w/o SGDF & 0.4661 & 0.3456  & 0.4877 \\
                             &  w/o CL & 0.5772 & 0.5408 & 0.6176 \\
                             & GDCN  & \textbf{ 0.6335} & \textbf{0.5638 } & \textbf{0.6522}  \\
\bottomrule
\end{tabular}
\label{tab:Ablation}
\end{table}

\begin{figure}[htbp]
\centering
\subfigure[] { 
\includegraphics[width=0.7\columnwidth]{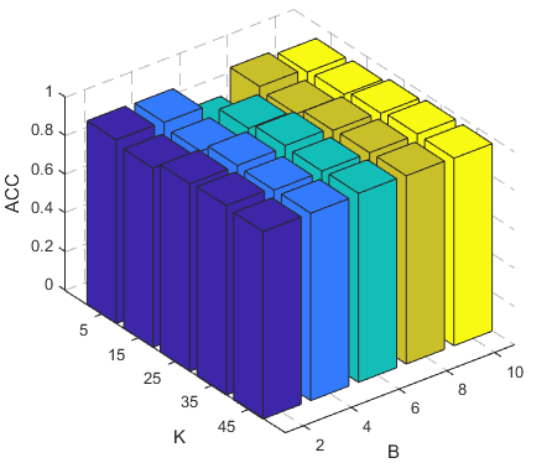}
}\hspace{0.25cm}
\subfigure[] { 
\includegraphics[width=0.7\columnwidth]{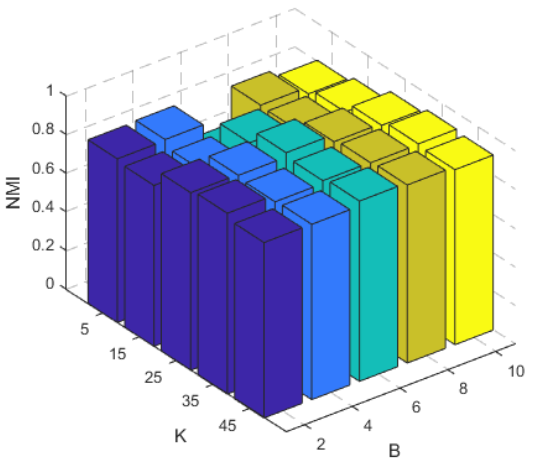}
}
\caption{The hyper-parameter analysis on NGs.}
\label{fig:hyper} 
\end{figure}

\textbf{Effectiveness of SGDF module.} The fused representation is set to $c_i$, which is the concatenation of all view-specific representations. The ``w/o SGDF" denotes the removal of the SGDF module from GDCN. Table \ref{tab:Ablation} illustrates that, in the ACC term, the results of ``w/o SGDF" are $11.80$, $1.17$, $4.81$, and $16.74$ percent less than those of GDCN on NGs, Synthetic3d, Caltech5V, and Wikipedia, respectively. SGDF implements the robust fusion of multi-view data by utilizing a multiple generative mechanism to address the low-quality issue. The results demonstrate that SGDF significantly improves multi-view clustering performance.

\textbf{Validity of CL module.} The ``w/o CL" represents the elimination of the Contrastive Learning module from GDCN. According to Table \ref{tab:Ablation}, the results of ``w/o CL" are lower than those of GDCN by $36.20$, $21.17$, $19.42$, and $5.63$ percent in ACC term on NGs, Synthetic3d, Caltech5V, and Wikipedia. The Contrastive Learning module can obtain the consistency of multi-view data. Therefore, contrastive learning enhances the results of deep multi-view clustering tasks.

\subsection{Convergence, Visualization, and Parameter Analysis.} 
As illustrated in Fig. \ref{fig:Visualization and convergence} (a), the value of ACC first increases in the early stage (i.e., the first 40 epochs) and then fluctuates in a narrow range. When epoch equals 200, there is a sudden change in ACC due to the transition from the pre-training phase to the fine-tuning phase. These results all confirm the convergence of GDCN. In addition, to further verify the effectiveness of the proposed GDCN, we visualize the common representations $\{\hat{h}_i\}_{i=1}^{N}$ after convergence by the t-SNE method \cite{van2008visualizing} in Fig. \ref{fig:Visualization and convergence} (b). The NGs dataset is divided into $5$ categories with clear cluster boundaries. Fig. \ref{fig:hyper} shows the analysis experiment of two hyperparameters, sampling times $B$ and discrete points $K$. The range of $B$ is $2$ to $10$. The value of $K$ ranges from $5$ to $45$. It can be concluded that ACC and NMI are insensitive to the two hyperparameters. 

\section{Conclusion and Future Work}
This paper proposes a novel Stochastic Generative Diffusion Fusion (SGDF) method to address the problem of low-quality data. Based on the SGDF module, we present the Generative Diffusion Contrastive Network (GDCN). It achieves the state-of-the-art results in Multi-View Clustering (MVC) tasks. We believe that GDCN has a very promising application prospect.

\newpage

\bibliographystyle{IEEE}
\bibliography{SPL}

\end{document}